\documentclass[letterpaper, 10 pt, conference]{ieeeconf}  

\IEEEoverridecommandlockouts                              

\usepackage{multirow}       
\usepackage{xcolor}         
\usepackage{booktabs}       
\usepackage{caption}        
\usepackage{graphicx} 
\usepackage{amsmath} 
\usepackage{cite}    
\makeatletter
\let\NAT@parse\undefined
\makeatother
\usepackage[colorlinks=true,linkcolor=blue,citecolor=blue]{hyperref}
\usepackage{array}
\usepackage{amssymb}  
\usepackage{makecell}
\usepackage{booktabs}
\usepackage{array}
\usepackage{tabularx}
\usepackage{colortbl}
\definecolor{myblue}{rgb}{.70,.96,.97}
\definecolor{mygrey}{rgb}{.97,.97,.97}
\title{\LARGE \bf
GLUE: Global-Local Unified Encoding for Imitation Learning via Key-Patch Tracking
}

\author{Ye Chen$^{1}$, Zichen Zhou$^{1}$, Jianyu Dou$^{1}$, Te Cui$^{1}$, Yi Yang$^{1}$, Yufeng Yue$^{1}$$^\dagger$
\\  $^{1}$Beijing Institute of Technology
}

\begin{document}

\maketitle
\thispagestyle{empty}
\pagestyle{empty}

\begin{abstract}
In recent years, visual representation learning has gained widespread attention in robotic imitation learning. However, in complex  Out-of-Distribution(OOD) settings characterized by clutter and occlusion, the attention of global visual representations can be diluted or interfered, leading to degraded policy performance. The invariance of local representations for task-relevant objects offers a solution. By efficiently utilizing these local representations, training and testing data can be mapped to a more similar feature space, thereby mitigating the covariate shift problem. Accordingly, we propose GLUE, a global-local unified encoding framework for imitation learning based on key-patch tracking. GLUE selects and tracks key-patches as critical local representations by employing a text-guided mechanism. It features a novel fusion framework where global patch features query local patches to distill essential information, yielding fine-grained local features with low heterogeneity relative to the global context. This fused representation steers the robot’s visual attention toward task-relevant objects and preserves precise global context, which together align the training and testing distributions into a similar and task-informative feature space, ultimately enhancing the robustness of the imitation learning policy. Experiments demonstrate that GLUE achieves strong performance across diverse tasks in both simulation and real-world settings, outperforming the strongest baseline by 17.6\% in simulation, 36.3\% in real-world environments, and 58.3\% on real-world generalization settings. The project website of GLUE is available at \href{https://GLUE666.github.io/}{https://GLUE666.github.io/}.

\end{abstract}

\section{INTRODUCTION}

 Imitation learning from demonstrations has emerged as a dominant paradigm for robot behavior learning \cite{dp,act,bcz,vip}. A key long-term goal is to enable a manipulation policy to address the covariate shift problem in Out-of-Distribution (OOD) scenarios characterized by clutter and occlusion without increasing the amount of training data. To achieve this goal, a key precondition is to provide the robot’s policy with robust visual conditioning that enables both accurate understanding of the overall scene context and precise attention to task-relevant regions. This can map the original data distribution and the actual test distribution to a similar and task-informative feature space, thereby mitigating the covariate shift.

\begin{figure}[t]
    \centering
    \includegraphics[height=0.37\textwidth]{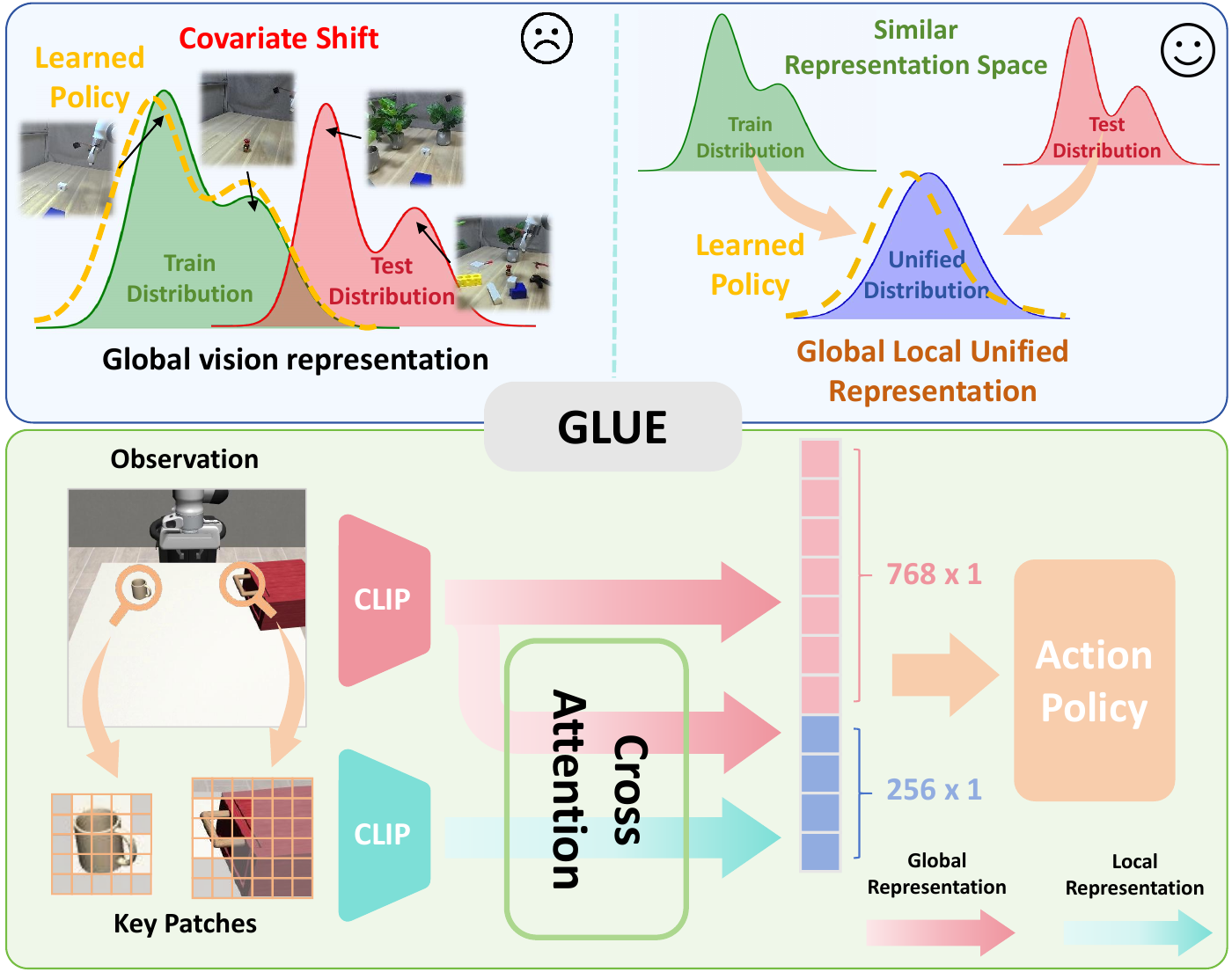}
    
    \caption{We propose GLUE (\textbf{G}lobal–\textbf{L}ocal \textbf{U}nified \textbf{E}ncoding), an encoding framework for imitation learning via key-patch tracking. GLUE fuses global and local information to align  test and training distributions in a similar, task-informative feature space, mitigating the covariate shift problem.}
    
    \label{fig:1}
    \vspace{-0.6cm}
\end{figure}

Currently, the mainstream visual encoding frameworks in imitation learning tend to  compress the entire scene's visual information into a single, global feature \cite{dp,act,openvla,rt2,octo}. As shown in Fig. \ref{fig:1}, this end-to-end encoding paradigm achieves strong fitting performance in in-domain settings where the distribution resembles the training data, successfully handling a wide range of complex tasks. However, real-world scenarios usually involve significant clutter and occlusion, resulting in OOD conditions that differ substantially from the training distribution. At this point, a single global representation can be easily can be easily diluted by task-irrelevant objects or disrupted by object occlusions. This is catastrophic for models trained with homogenous data.

A straightforward approach to addressing this problem is to introduce object-centric representations to strengthen the perception of key objects which exhibits good invariance during operation. This direction has garnered significant attention and demonstrated considerable potential\cite{viola, groot, kalm, spot, controlvla, p3po}. However, as scene complexity increases, the discrepancy between training and testing distributions is further amplified, while restricting the visual input to RGB observations further limits the model’s ability to cope with the covariate shift problem, resulting in a severe degradation of existing methods’ performance. Upon a closer examination of existing object-centric encoding approaches, we highlight two core challenges in such scenarios.

The first challenge is, \textbf{how to efficiently and adaptively define and track key regions in RGB observations.} The mainstream region cues in existing object-centric encoding methods can be categorized into three types: bounding boxes \cite{viola, devin2018deep, migimatsu2020object, mandlekar2020learning}, object poses \cite{spot, pan2025omnimanip, tremblay2018deep}, and keypoints \cite{p3po, skil, rekep, abc}. Bounding boxes and object poses provide coarse instance-level cues, where mislocalization or occlusion of a single representation can heavily impact model decisions and reduce robustness. Keypoints provide pixel-level cues, being suitable for 3D coordinates while limiting the information captured from RGB input. In contrast, key-patches are of moderate granularity, requiring only low tracking cost while maintaining robustness. They are also highly compatible with the patch-level encoding scheme of mainstream vision transformer\cite{vit} architectures. Based on the above considerations, GLUE selects key-patches as cues for critical regions. Simultaneously, existing methods locate key regions either with excessive human guidance to select from numerous masks\cite{groot, wen2024object, manipllm,prismdp} or via unstable self-supervised approaches with insufficient accuracy \cite{viola, chapin2025object,composing}. To address this, GLUE introduces a task-driven, text-guided framework for automatic annotation and real-time tracking of key regions based on off-the-shelf grounding and tracking models \cite{grounding,SAM,cotracker3}, significantly reducing human involvement.

The second challenge is, \textbf{how to align cluttered and occluded test distributions with in-domain training distributions in a task-informative feature space.} Most object-centric encoding methods overfocus on local regions, which may align training and testing distributions in a similar feature space, but the suppression of scene context and object–environment relations leaves the features insufficient for guiding downstream action generation\cite{p3po, skil, spot}.
Based on this, some approaches attempt to combine global and local features , yet most still encode them independently, leading to heterogeneous representations that hinder global–local unification \cite{zhicheng2025object}. To address this, GLUE introduces a global–local patch feature fusion framework based on a pretrained vision transformer encoder and a multi-head cross-attention network, enhancing interaction between global and local feature encoding. Specifically, GLUE separately encode the full image and task-relevant patches with CLIP to obtain global and local grid features. These features are then fused via a multi-head cross-attention network, where the global features act as queries to distill information from the key local regions. The resulting refined local feature is then concatenated with the CLS token from CLIP’s image-level encoding to form a comprehensive representation, mapping training and test distributions into a similar, task-informative feature space.

In summary, we introduce GLUE, a global–local unified encoding framework for  imitation learning based on key-patch tracking. As illustrated in Fig. \ref{fig:1}, GLUE leverages effective interaction between global and local information to map OOD test scene distributions and in-domain training distributions into a similar and task-informative feature space, effectively mitigating the covariate shift problem. The contributions of GLUE can be summarized as follows:

(1) GLUE selects moderately granular key-patches that are highly compatible with RGB observations as critical-region cues and implements a task-driven, text-guided framework for automatic annotation and real-time tracking.

(2) GLUE designs an efficient global–local fusion framework, mitigating heterogeneity between global and local encodings, thus mapping training and test distributions into a similar, task-informative feature space.

(3) GLUE demonstrates strong performance across both simulation and real-world tasks, surpassing the strongest baseline by 17.6\% in  simulation, 36.3\% on  real-world tasks, and 58.3\% on real-world generalization settings.


\section{RELATED WORKS}

\subsection{Vision Representation for Manipulation Policy}

Visual representations are foundational for translating pixels into actionable knowledge for robotic manipulation. Early end-to-end visuomotor policies \cite{8461196, finn2016deep, florence2019self, levine2016end}, which typically used CNN encoders trained from scratch, struggled with generalization due to limited in-domain data. 

The advent of visual models pre-trained on large-scale such as ImageNet \cite{imagenet} offered a potent solution to these challenges. By leveraging rich semantic priors from these datasets, these models significantly boosted both data efficiency and generalization, quickly becoming the standard paradigm \cite{act, dp, ibc, rt1}. Nonetheless, as features from classification-centric data failed to capture the visual priors necessary for manipulation, the field shifted to leveraging foundation models like CLIP and DINO \cite{openvla, octo, rt2}. Subsequently, models pre-trained on large-scale video datasets further revolutionized the field by enabling the learning of dynamic visual representations \cite{r3m,mvp,vip}. Despite their significant progress, these methods rely on a single global vector to compress all task-relevant information, often at the cost of losing the crucial spatial details required for fine-grained manipulation.

In this work, we propose GLUE, which directly addresses this information bottleneck by explicitly extracting and fusing task-relevant object features with the global scene context, leading to more precise and robust manipulation policies under complex scenarios.

\subsection{Object-centric Representation Learning}

Object-centric representation shifts the focus from holistic scene encoding to learning distinct, independent features for each object. The pursuit of efficient and robust object-centric representations has long been a central theme in the field of robotics. The most common and straightforward forms of object-centric representation are bounding boxes \cite{viola, devin2018deep, migimatsu2020object, mandlekar2020learning} and object poses \cite{spot, pan2025omnimanip, tremblay2018deep}. These types of representation methods are very coarse and extremely fragile in complex, dynamic environments. Another popular method is to use keypoints to represent the object's position or manipulation points \cite{p3po, skil, rekep, abc}. Although keypoint-based models are highly efficient, they excessively compress visual information, leading to the loss of a significant amount of useful information. Other methods use point clouds as an object-centric representation \cite{groot, cordvip}, but these representations remain too sparse and are more difficult to track accurately.

In contrast, GLUE's strategy of extracting image patches centered on tracked object keypoints offers a dual advantage: it ensures robust tracking in dynamic settings while simultaneously retaining high-fidelity local visual features, directly mitigating the drawbacks of earlier methods.

\section{METHOD}

\subsection{Framework Overview} 

In cluttered and occluded OOD scenarios with only RGB information available, our goal is to efficiently localize and track task-relevant patches in real time, thereby obtaining representations that preserve global understanding while maintaining focus on critical regions, thus mitigating the covariate shift problem.
An overview of our method is illustrated in Fig. \ref{fig:3}. First, we design a one-time keypoints extracting framework to extract task-relevant points from the initial image, requiring only low-cost task-relevant text prompts. Next, an off-the-shelf tracking model\cite{cotracker3} is employed for real-time tracking of keypoints without extra effort. The tracked keypoints are used to locate critical patches within the fine-grained grid, providing region cues for subsequent encoding. Finally, we design a unified global-local encoding framework with two CLIP networks\cite{clip} and a multi-head cross-attention network, providing the policy model\cite{dp} with powerful visual conditioning information. The detailed framework design will be presented in the following sections.

\subsection{Key Points Localization and Real-Time Tracking }
We adopt key-patch as a moderately granular cue for critical regions. To reduce localization and tracking costs, each key-patch is anchored to a keypoint, with localization and real-time tracking performed at the keypoint level. In this module, we first extract object keypoints in the initial frame based on task-related text queries, with the help of a vision foundation model\cite{clip} and pretrained vision models\cite{dinov2,SAM}. As visual observations update, we track the keypoints in real time using an off-the-shelf tracker initialized from the first-frame queries\cite{cotracker3}, providing the basis for subsequent key-patch localization. The overall pipeline is illustrated in Fig. \ref{fig:3}.

\begin{figure}[t]
    \centering
    \includegraphics[width=0.48\textwidth]{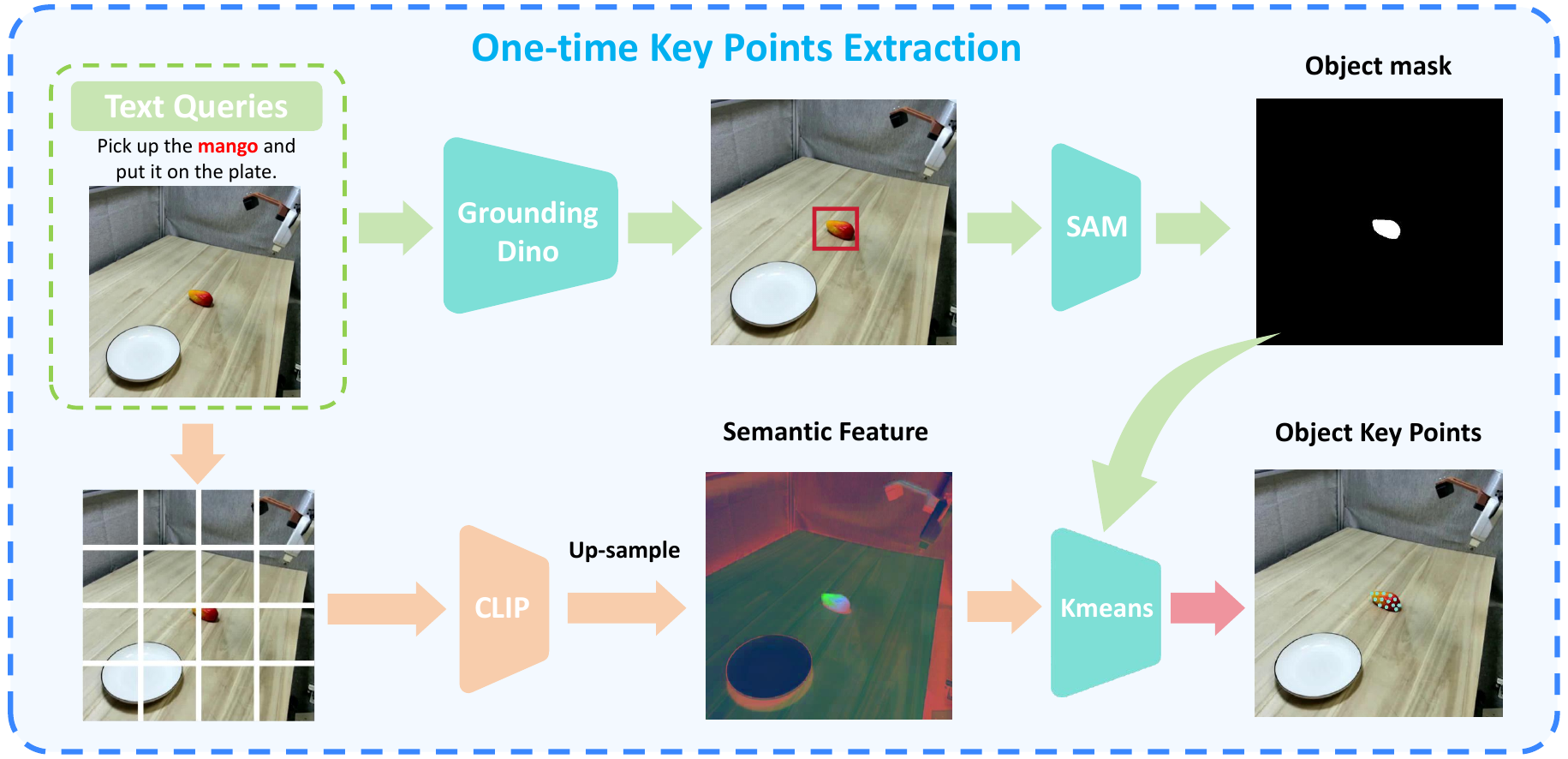}
    \caption{One-time keypoint extraction framework: (1) Given the initial frame, a single task-specific text prompt is fed to Grounding DINO\cite{grounding}, which guides SAM\cite{SAM} to obtain object-related masks. (2) The pre-segmented image is encoded by CLIP\cite{clip}, and the upsampled features are concatenated to produce a per-pixel feature map. (3) Keypoints are extracted via per-pixel feature clustering within the object masks.}
    \label{fig:2}
    \vspace{-4pt}
\end{figure}

\textbf{One-time Key Points Extraction:} For each task, a single task-specific text query is provided once to automatically generate initial keypoints from the first frame, which then serve as the basis for subsequent keypoint tracking. As illustrated in Fig. \ref{fig:2}, the one-time keypoint extraction process consists of three stages: text-guided mask generation,  pre-segmentation based semantic mapping, and keypoint selection.

1) Text-guided Mask Generation: For each specific task, we begin by providing  one-time task-relevant text queries ${\{{c}_i\}}_{i=1}^{N_{obj}}$. Given the initial frame $I_1 \in R^{H\times W\times 3}$ before each task execution, we first feed $I_1$ and ${\{{c}_i\}}_{i=1}^{N_{obj}}$ into a text-guided bounding box generation model $Det(\cdot , \cdot)$ (Grounding DINO\cite{grounding}) to obtain a set of task-relevant bounding boxes ${\{{B}_i\}}_{i=1}^{N_{obj}}$. These bounding boxes are then used as prompts and fed along with $I_1$ into a mask generation model $Seg(\cdot , \cdot)$ (SAM\cite{SAM}), producing task-relevant object masks ${\{{M}_i\}}_{i=1}^{N_{obj}}$. The entire process is summarized as follows:
\begin{equation}
{\{{M}_i\}}_{i=1}^{N_{obj}}=Seg( I_1 , Det(I_1, {\{{c}_i\}}_{i=1}^{N_{obj}}))
\end{equation}

2) Pre-segmentation Based Semantic Mapping: To obtain higher-resolution per-pixel semantic features, we first pre-segment $I_1$ into a fine-grained $N_{grid}\times N_{grid}$ grid before feature encoding, resulting in a set of sub-images  ${\{{I}_j\in R^{h\times w \times 3}\}}_{j=1}^{N_{grid}^{2}}$. For each sub-image, features are encoded using a vision transformer\cite{vit} based vision foundation model (e.g., CLIP\cite{clip} and DINO\cite{dinov2}), yielding a low-resolution patch-wise feature map ${\{{f}_i^{'}\in R^{n\times n \times N_f}\}}_{i=1}^{N_{grid}^{2}}$ that is subsequently upsampled via bilinear interpolation to obtain a full-resolution per-pixel semantic feature map  ${\{{f}_i\in R^{h\times w \times N_f}\}}_{i=1}^{N_{grid}^{2}}$ ,where $n$ denotes the number of grid cells partitioned by the foundation model, and $N_f$ represents the dimensionality of the feature vectors. By concatenating  ${\{{f}_i\}}_{i=1}^{N_{grid}^{2}}$, we finally reconstruct the full-resolution per-pixel feature map $\mathcal{F}\in R^{H \times W \times N_f}$.

3) Keypoint Selection: Using  ${\{{M}_i\}}_{i=1}^{N_{obj}}$, we perform object-region segmentation on  $\mathcal{F}$ to obtain a set of object-centric feature maps  ${\{ \mathcal{F}[{M}_i] \}}_{i=1}^{N_{obj}}$, and apply K-means to cluster these maps into 
$N_{cluster}$ clusters, selecting the point closest to each cluster center as the corresponding keypoint.

\textbf{Real-Time Key Points Tracking:} Given a new RGB observation $I_{t+1}$, we feed the full RGB observation sequence ${\{{I}_m}\}_{m=1}^{t+1}$ along with the initial keypoint queries ${\{(h_1^i,w_1^i,o_1^i)\}}_{i=1}^{N_{cluster}}$ into an off-the-shelf tracker\cite{cotracker3} to obtain real-time tracked keypoints  ${\{(h_{t+1}^i,w_{t+1}^i,o_{t+1}^i)\}}_{i=1}^{N_{cluster}}$, where $o_t^i$ indicates the visibility of the $i$-th keypoint at time $t$. We define the whole process as:
\begin{subequations}\label{eq:action}
\begin{align}
{\{p_{t}^{i}\}}_{i=1}^{N_{cluster}}  :&= {(h_t^i,w_t^i,o_t^i)\}}_{i=1}^{N_{cluster}}\\
{\{p_{t+1}^{i}\}}_{i=1}^{N_{cluster}} &=Track({\{{I}_m}\}_{m=1}^{t+1},{\{p_1^i\}}_{i=1}^{N_{cluster}}) 
\end{align}
\end{subequations}


\subsection{Global-Local Unified  Encoding and Action Generation
} \label{Section.C}

\begin{figure*}[t]
    \centering
    \includegraphics[width=1.032\textwidth]{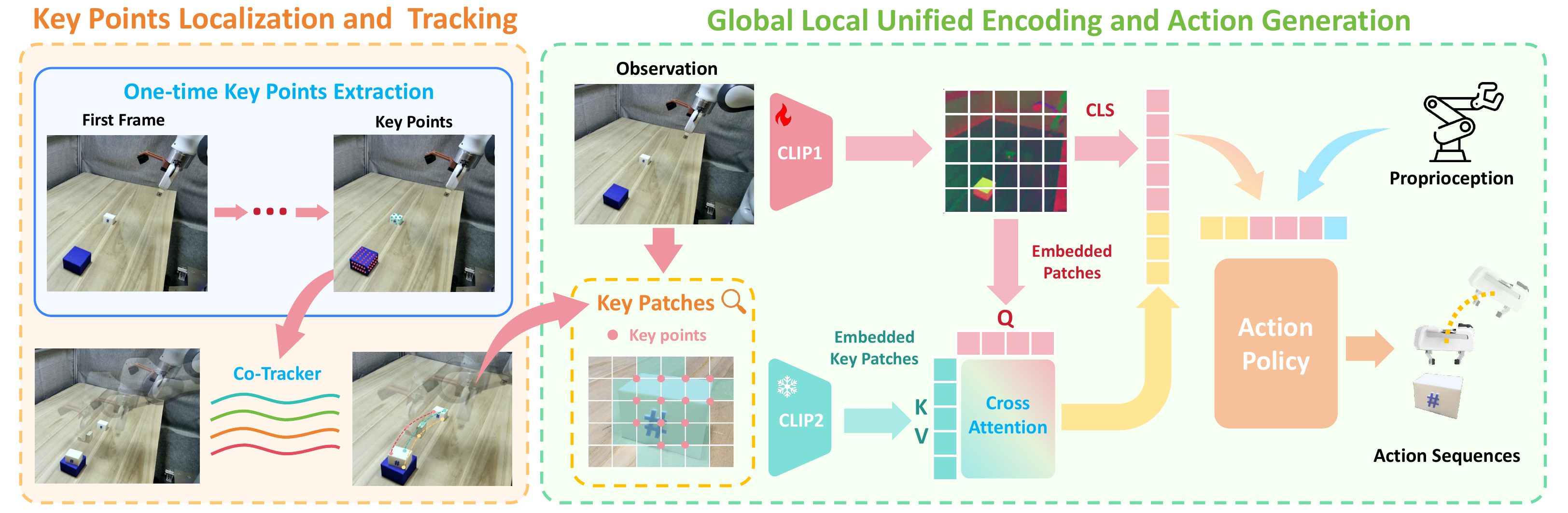}
    \caption{Overall framework of GLUE. GLUE first extracts keypoints from the initial frame and tracks them to localize key patches. Both key patches and the full image are encoded with CLIP to obtain local and global features, which are fused via a multi-head cross-attention network and combined with the CLS token to provide unified global–local visual condition for the action policy.}
    \label{fig:3}
    \vspace{-5pt}
\end{figure*}

In this module, we first encode the RGB observation with a learnable CLIP network\cite{clip} to obtain the global CLS token and global grid features. The tracked keypoints are then used to identify the corresponding key-patches, which are encoded via a frozen CLIP network\cite{clip} to produce local grid features. Global grid features are further used to distill fine-grained local information from the key grids, and the resulting features are concatenated with the global CLS token and robot-state information to form a conditional encoding. Finally, the conditional encoding is fed into a diffusion policy head\cite{dp} to predict actions in real time from noise. The overall pipeline is illustrated in Fig. \ref{fig:3}.

\textbf{Global-Local Unified Encoding:}  Given  $I_{t+1}$ and ${\{(h_{t+1}^i,w_{t+1}^i,o_{t+1}^i)\}}_{i=1}^{N_{cluster}}$, we first encode the visual information to obtain a robust global–local unified representation. As illustrated in Fig. 3, the overall encoding process can be divided into three stages: global grid feature encoding, local grid feature encoding, and distillation of fine-grained local feature.

1) global grid feature encoding: In this stage, we first encode $I_{t+1}$  using a learnable CLIP network\cite{clip}. To preserve patch-level features, the pooling layer of this CLIP network is removed. After encoding $I_{t+1}$, we obtain global features ${\mathcal{F}}_{global}\in R^{197 \times 768}$consisting of 196 patch features and one CLS token. ${\mathcal{F}}_{global}$ is then separated into global patch features ${\mathcal{F}}_{patch1}\in R^{196 \times 768}$ and the global CLS token ${\mathcal{F}}_{CLS}\in R^{768}$.

2) local grid feature encoding: Leveraging the key-point information ${(h_{t+1}^i,w_{t+1}^i,o_{t+1}^i)\}}_{i=1}^{N_{cluster}}$, we first perform localization of the key-patch regions. Specifically, $I_{t+1}$ is divided into a $28\times28$ grid, and the key-patch coordinates are determined based on the grid cells containing the keypoints. The key patches inherit the visibility information of the corresponding keypoints, resulting in a complete set of key patch representations ${(x_{t+1}^i,y_{t+1}^i,o_{t+1}^i)\}}_{i=1}^{N_{cluster}}$. Next, $I_1$ is pre-segmented into a grid to obtain sub-images ${\{I_k\in R^{\frac{h}{2} \times \frac{w}{2} \times 3}\}_{k=1}^{4}}$, each of which is encoded using a frozen CLIP network\cite{clip} with the CLS token removed to produce sub-feature vectors ${\{{\mathcal{F}}_{local}^{j}\in R^{196 \times 768}\}}_{j=1}^{4}$. These vectors are then restored and concatenated to form the grid feature map ${\mathcal{M}}_{local}\in R^{28 \times 28 \times 768}$. Finally, we use ${(x_{t+1}^i,y_{t+1}^i)\}}_{i=1}^{N_{cluster}}$ to select key-patch features within ${\mathcal{M}}_{local}$ and zero out the features of occluded patches, yielding the final local grid feature ${\mathcal{F}}_{patch2}\in R^{N_{cluster} \times 768}$.

3) distillation of fine-grained local feature: To mitigate the heterogeneity between local and global features, we distill fine-grained local features via a multi-head cross-attention network $MHA(\cdot , \cdot)$ . Specifically, we add positional encoding to ${\mathcal{F}}_{patch1}$ and ${\mathcal{F}}_{patch2}$, projecting ${\mathcal{F}}_{patch1}$ into the query matrix $Q$ and ${\mathcal{F}}_{patch2}$ into the key and value matrices $K$ and $V$ with the help of three learnable projection matrices. The multi-head cross-attention network then encodes these into intermediate features  $\tilde{{O}} \in R^{196 \times 768}$, which are subsequently processed through a linear mapping and average pooling to obtain the fine-grained local feature ${\mathcal{F}}^{'}\in R^{256}$  distilled from the local patches by querying with the global patch features. The overall process can be formally defined as:
\begin{equation}
{\mathcal{F}}^{'}=Pool(Proj(MHA({{\mathcal{F}}_{patch1},\mathcal{F}}_{patch2},{\mathcal{F}}_{patch2})))
\end{equation}

\textbf{Action Generation:} To obtain a global–local unified visual condition, we concatenate ${\mathcal{F}}^{'}$ with ${\mathcal{F}}_{CLS}$ and further fuse it with the robot joint state ${\mathcal{S}}\in R^{7}$, yielding the final conditional representation ${\mathcal{W}}\in R^{1031}$. Conditioned on the global–local unified representation ${\mathcal{W}}$, we employ a diffusion-based action head to predict the robot’s future action sequence. Following Diffusion Policy\cite{dp}, we adopt a CNN-based U-Net as the noise prediction network. Given a single observation $o_t$, the model predicts an action chunk ${A}_t = \{a_t, a_{t+1}, \dots, a_{t+H}\}$. Starting with a fully noisy action chunk $A^{k}_t$ sampled from Gaussian noise, the denoising network $\epsilon_\theta$ performs $k$ denoising iterations to gradually produce a final action chunk $A^{k}_t$, which is then executed by the robot arm: 
\begin{equation}
A^{k-1}_t = \alpha_k(A^{k}_t -\gamma_k\epsilon_\theta(\mathcal{W},A^{k}_t,k) + \mathcal{N}(0, \sigma_k^2 I))
\end{equation}
where $\mathcal{N}(0, \sigma_k^2 I)$ is a gaussian noise, and the choice of $\alpha_k,\gamma_k,\sigma_k$ is depending on $k$.

During training process, We employ DDIM\cite{ddim} for accelerated sampling. The training objective is to predict the noise added to the original data based on the following loss function:
\begin{equation}
    \mathcal{L}= MSE(\epsilon^k, \epsilon_\theta(\bar{\alpha_k}A^{0}_t+\bar{\beta}\epsilon^k, \mathcal{W}, k))
\end{equation}

\begin{table*}[th]
\caption{
    MimicGen simulation experiment results. We report the average success rate of each method across three random seeds for all eight tasks. GLUE achieves the highest success rate on seven tasks, as well as the highest average success rate across all tasks.
}
\centering
\small 
\renewcommand\arraystretch{1.3}
\setlength{\tabcolsep}{0mm}
\begin{tabularx}{\linewidth}{l *{9}{>{\centering\arraybackslash}X}}
\toprule
\multirow{2}{*}{\footnotesize Method} &
\begin{tabular}[t]{@{}c@{}} \footnotesize Stack \\ \footnotesize D1\end{tabular} &
\begin{tabular}[t]{@{}c@{}} \footnotesize Stack Three \\ \footnotesize D1\end{tabular} &
\begin{tabular}[t]{@{}c@{}} \footnotesize Mug Cleanup \\ \footnotesize D1\end{tabular} &
\begin{tabular}[t]{@{}c@{}} \footnotesize Threading \\ \footnotesize D0\end{tabular} &
\begin{tabular}[t]{@{}c@{}} \footnotesize Square \\ \footnotesize D0\end{tabular} &
\begin{tabular}[t]{@{}c@{}} \footnotesize Kitchen \\ \footnotesize D0\end{tabular} &
\begin{tabular}[t]{@{}c@{}} \footnotesize Nut Assembly \\ \footnotesize D0\end{tabular} &
\begin{tabular}[t]{@{}c@{}} \footnotesize Coffee \\ \footnotesize D1\end{tabular} &
\multirow{2}{*}{\footnotesize Average}\\
\midrule

 DP \cite{dp}  & 26.3 & 0.0 & 0.0 & 31.0 & 7.7 & 85.3 & 0.0 & 0.0 & 18.8 \\
\rowcolor{mygrey} DP3 \cite{dp3}        & 24.3 & 0.0 & 15.0 & 48.7 & \textbf{51.0} & \textbf{97.0} & 5.0 & 34.3 & 34.4 \\
 ACT \cite{act}       & 51.3 & 25.7 & 28.7 & 23.0 & 27.0 & 96.0 & 3.7 & 22.7 & 34.8 \\

\midrule
    \noalign{\vskip -0.5ex}
    \rowcolor{mygrey} GLUE        & \textbf{76.0} & \textbf{67.7} & \textbf{33.0} & \textbf{52.3} & 43.0 & \textbf{97.0} & \textbf{12.7} & \textbf{37.3} & \textbf{52.4} \\
\bottomrule
\end{tabularx}

\label{tab:table1}
\end{table*}

\section{Simulation Experiment}
\subsection{Experimental Setup}

\textbf{Simulation Benchmark:} We conduct our simulation experiments in the MimicGen\cite{mimicgen} environment, which is built on the Mujoco\cite{mujoco} simulator. MimicGen leverages a small number of human demonstrations to perform data generation, producing multi-modal, non-Markovian trajectories with strong generalization across object configurations. Moreover, it encompasses diverse tasks that require high-precision manipulation and long-horizon reasoning, making it a powerful benchmark for evaluating imitation learning methods.

\begin{figure}[t]
    \centering
    \includegraphics[width=0.48\textwidth]{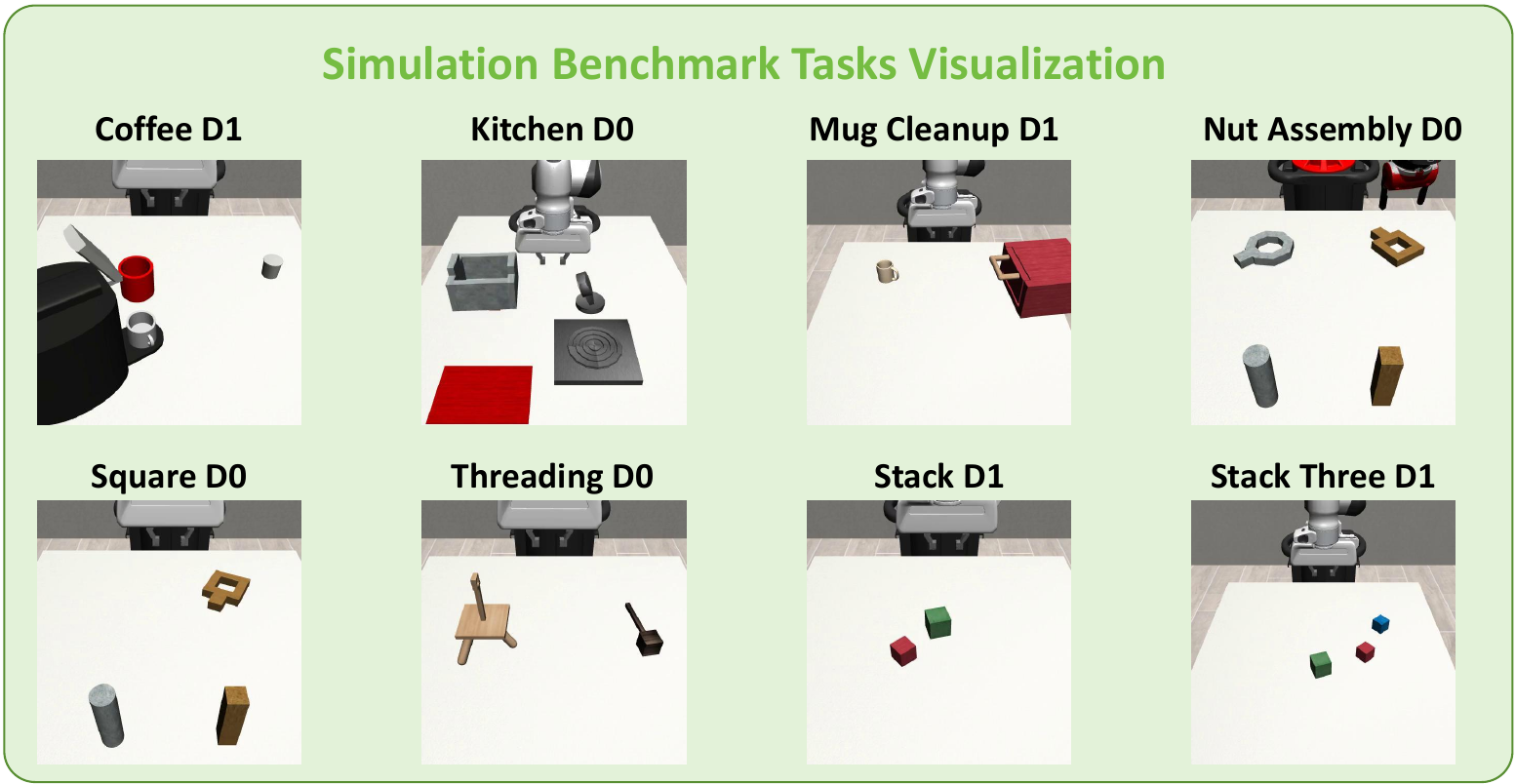}
    
    \caption{We select eight tasks from the MimicGen benchmark for our simulation experiments: Coffee D1, Kitchen D0, Mug Cleanup D1, Nut Assembly D0, Square D0, Threading D0, Stack D1, and Stack Three D1. }
    
    \label{fig:4}
    \vspace{-5pt}
\end{figure}

\textbf{Evaluation Scenes:}  To comprehensively evaluate model performance in simulation, we select eight tasks from the MimicGen\cite{mimicgen} benchmark, covering high object-position generalization (Stack D1, Stack Three D1, Mug Cleanup D1), fine-grained manipulation (Square D0, Threading D0, Coffee D1), long-horizon reasoning (Kitchen D0), and tasks with novel robotic arms (Nut Assembly D0). Task visualizations are shown in Fig. \ref{fig:4}.

\textbf{Comparison Baseline:} We compare our method with three state-of-the-art imitation learning algorithms: Diffusion Policy (DP) \cite{dp}, Action Chunk Transformer (ACT) \cite{act}, 3D Diffusion Policy (DP3) \cite{dp3}. Wherein DP and ACT are 2D policies, and DP3 is a 3D policy. For 2D policies, we use a $512\times 512$ third-person RGB image as visual observations, while for 3D policies, we construct point clouds by projecting a $512\times 512$ third-person depth image. The point clouds are cropped to retain only the robot arm and relevant objects.

\textbf{Evaluation Metric:}  To evaluate model performance, we train on 100 demonstration trajectories.  Each model is trained for 500 epochs, with evaluations conducted every 100 epochs on 100 novel randomized object configurations. We report the results from the epoch achieving the highest success rate, averaged over three random seeds.

\subsection{Experiment Results}

The simulation results on MimicGen\cite{mimicgen} benchmark are summarized in Table \ref{tab:table1}. Across the eight simulation tasks, GLUE achieves the highest success rate on seven tasks and the highest average success rate overall, surpassing the strongest baseline by 17.6\%, which demonstrates its strong performance and robustness in simulation. In particular, on the high object-position generalization tasks Stack D1 and Stack Three D1, GLUE surpasses the strongest baseline by 24.7\% and 42\%, highlighting its robustness in handling diverse object configurations. These results indicate that, in in-domain scenarios, the global–local unified visual conditon provided by GLUE effectively reinforce guidance toward task-critical regions, further improving model performance.

\subsection{Ablation Study}
We perform ablation studies in simulation to examine the core module designs of GLUE. The effectiveness of these design choices is evaluated based on success rates across the simulation tasks.

\begin{table}[th]
\caption{
    Ablation results on local fine-grained feature across four simulation tasks.
GLUE outperforms GLUE-S on all tasks, exceeding it by 5.5\% on average success rate.
}
\centering
\small 
\renewcommand\arraystretch{1.3}
\setlength{\tabcolsep}{0mm}
\begin{tabularx}{\linewidth}{l *{9}{>{\centering\arraybackslash}X}}
\toprule
\multirow{2}{*}{\footnotesize Method} &
\begin{tabular}[t]{@{}c@{}} \footnotesize Stack \\D1\end{tabular} &
\begin{tabular}[t]{@{}c@{}} \footnotesize  Stack Three \\D1\end{tabular} &
\begin{tabular}[t]{@{}c@{}} \footnotesize  Mug Cleanup\\ D1\end{tabular} &
\begin{tabular}[t]{@{}c@{}} \footnotesize  Threading\\ D0\end{tabular} &
\multirow{2}{*}{\footnotesize Average}
\\
\midrule

GLUE-S  &67.4 & 60.7 &25.0&47.0&50.0 \\
\rowcolor{mygrey}GLUE     &\textbf{ 74.5} &\textbf{67.4} &\textbf{29.5}&\textbf{50.7} &\textbf{55.5} \\
\bottomrule
\end{tabularx}

\label{tab:ablation1}
\end{table}

\textbf{Ablation on local fine-grained feature:} We evaluate the contribution of our local fine-grained feature to model performance across four simulation tasks. The evaluation metric is the mean of the two highest success rates on each task, averaged over three random seeds. The results are shown in Table \ref{tab:ablation1}, where GLUE-S denotes a variant of GLUE without the local fine-grained feature in its conditional encoding. GLUE outperforms GLUE-S on all four tasks, achieving an average success rate 5.5\% higher, demonstrating the effectiveness of the local fine-grained feature.

\begin{table}[th]
\caption{
    Ablation results on key-patch numbers across three simulation tasks.
}
\centering
\small 
\renewcommand\arraystretch{1.3}
\setlength{\tabcolsep}{0mm}
\begin{tabularx}{\linewidth}{l *{9}{>{\centering\arraybackslash}X}}
\toprule
\begin{tabular}[t]{@{}c@{}} \footnotesize Key-patch\\ Numbers\end{tabular} &
\begin{tabular}[t]{@{}c@{}} \footnotesize Stack \\D1\end{tabular} &
\begin{tabular}[t]{@{}c@{}} \footnotesize  Stack Three \\D1\end{tabular} &
\begin{tabular}[t]{@{}c@{}} \footnotesize  Mug Cleanup\\ D1\end{tabular} &
\multirow{2}{*}{\footnotesize Average}
\\
\midrule

10  &72.7 & 65.4 &28.2&55.4 \\
\rowcolor{mygrey}15      & 74.5 &\textbf{67.4} &29.5&57.1  \\
20     & \textbf{76.9} & 64.9 & \textbf{30.5}&\textbf{57.4}   \\
\bottomrule
\end{tabularx}

\label{tab:ablation2}
\end{table}

\textbf{Ablation on Key-patch Numbers:} We evaluate the impact of the number of key-patches on model performance across three simulation tasks, with each key-patch localized by its corresponding keypoint. The evaluation metric is the mean of the two highest success rates achieved on each task, averaged over three seeds. The results are presented in Table \ref{tab:ablation2}, indicating that GLUE’s performance remains stable when using 10–20 key-patches. Overall, increasing the number of key-patches tends to improve success rates, suggesting that with greater computational resources, GLUE can achieve higher performance by leveraging more key-patches.

\section{Real World Experiment}
\subsection{Experimental Setup}

\textbf{Scene Setup:} For real-world evaluation, we use a 7-DOF Franka Emika Panda robot with impedance control. Visual observations are captured using a single RealSense camera in a third-person view. Expert demonstrations for all real-world tasks are collected via a SpaceMouse, sampled at 10 Hz.

\begin{figure}[t]
    \centering
    \includegraphics[width=0.48\textwidth]{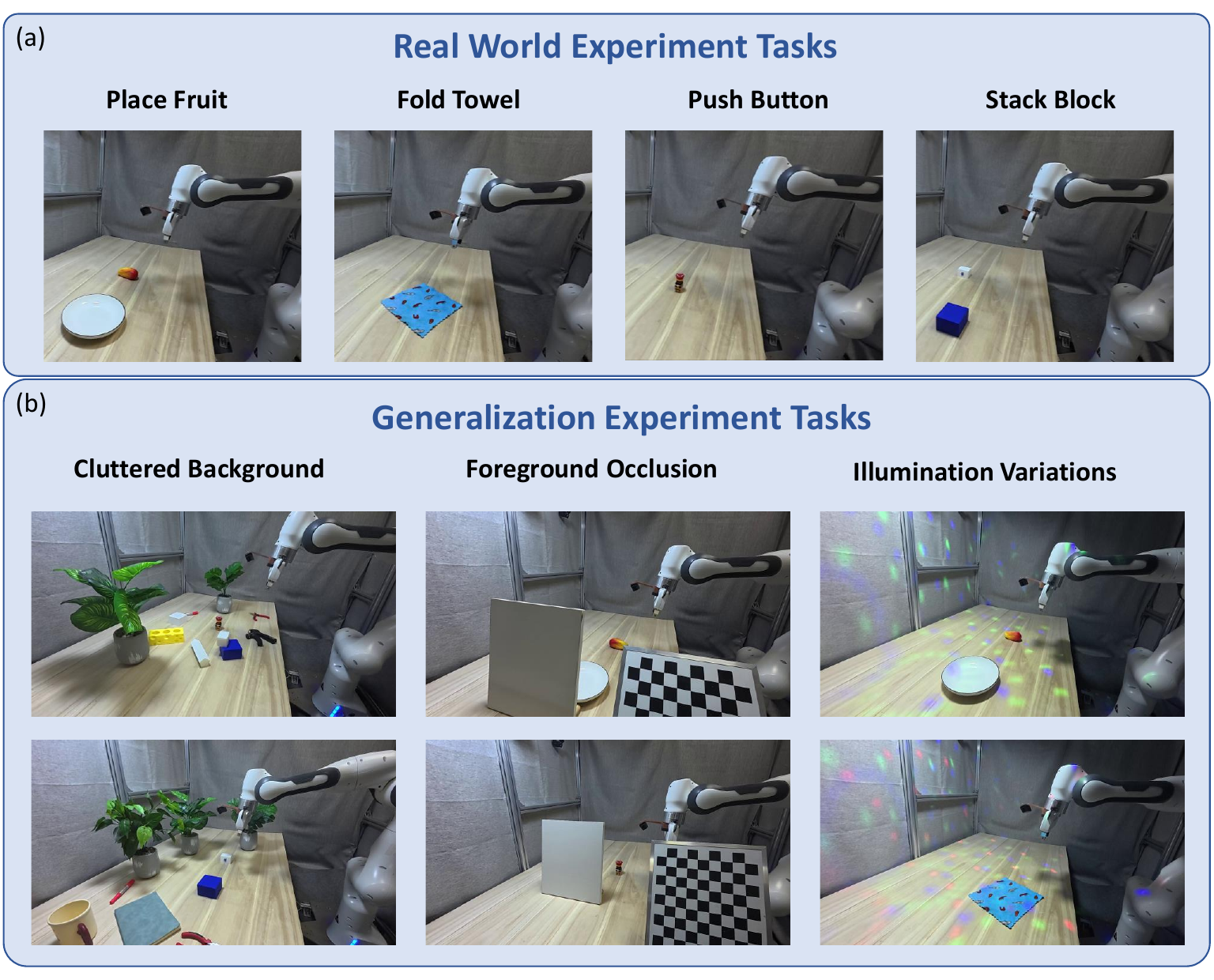}
    
    \caption{(a) We select four tasks for the real world experiment: Place Fruit, Fold Towel, Push Button, Stack Block. (b) We select three OOD scenario for generalization experiment: Cluttered Background, Foreground Occlusion, Illumination Variations}
    
    \label{fig:5}
    \vspace{-5pt}
\end{figure}

\textbf{Real World Tasks:} We conduct four experiments in real-world scenarios, briefly summarized as follows (Visualizations of all tasks  are shown in Fig.\ref{fig:5}(a):

1) Push Button: The robot closes its gripper and presses the red button on the table, with the position of button varied.

2) Stack Block: The robot stacks a white block on top of a blue block, with the positions of both blocks varied.

3) Place Fruit: The robot grasps a fruit and places it into a plate, with both the position and orientation of the fruit varied.

4) Fold Towel: The robot grasps the top-right corner of a towel and folds it, with both the position and rotation of the towel varied.

\textbf{Comparison Baseline:} For our real-world experiments, we select two 2D policies (DP\cite{dp} and ACT\cite{act}) as baselines. This ensures a fair comparison, as the GLUE benchmark operates exclusively on 2D visual representations. Additionally, we include GLUE-S in the evaluation to showcase the benefits of our fine-grained features in comparison with the baselines. For each method, we provide $512\times512$ resolution RGB images as visual observations.

\textbf{Evaluation Metric:}   In real-world experiments, we train each model using 50 demonstration trajectories per task. Each model is trained for 1,000 epochs and evaluated 20 times per task, with object positions varied to test generalization.
 And under the three generalization conditions, each model is evaluated 10 times per task.
\subsection{Experiment Results}

The real-world experimental results are shown in Table \ref{tab:real1}. GLUE achieves the highest success rate on all four tasks, outperforming the strongest baseline by 36.2\% on average and surpassing GLUE-S by 15\%. This demonstrates GLUE’s strong performance in real-world scenarios while further confirming the effectiveness of the local fine-grained feature in enhancing model performance. Supplementary video demonstrations
of the real-world experiments are available  at \href{https://GLUE666.github.io/}{https://GLUE666.github.io/}.

\begin{table}[th]
\caption{
    Real-world experimental results. GLUE achieves the highest success rates across all four tasks, demonstrating its strong performance in real-world settings.
}
\centering
\small 
\renewcommand\arraystretch{1.3}
\setlength{\tabcolsep}{0mm}
\begin{tabularx}{\linewidth}{l *{9}{>{\centering\arraybackslash}X}}
\toprule
\multirow{2}{*}{\footnotesize Method} &
\begin{tabular}[t]{@{}c@{}} \footnotesize Push \\ \footnotesize Button\end{tabular} &
\begin{tabular}[t]{@{}c@{}} \footnotesize Stack  \\ \footnotesize Block\end{tabular} &
\begin{tabular}[t]{@{}c@{}} \footnotesize Place \\ \footnotesize Fruit\end{tabular} &
\begin{tabular}[t]{@{}c@{}} \footnotesize Fold \\ \footnotesize Towel\end{tabular} &
\multirow{2}{*}{\footnotesize Average}\\
\midrule

 DP \cite{dp}  & 20.0 & 10.0 & 55.0 & 60.0 &36.2 \\
\rowcolor{mygrey}  ACT \cite{act}        & 55.0 & 50.0 & 30.0 & 60.0 & 48.8  \\
GLUE-S     & 70.0 & 55.0 & 80.0 & 75.0 & 70.0  \\

\midrule
    \noalign{\vskip -0.5ex}
    \rowcolor{mygrey} GLUE        & \textbf{80.0} & \textbf{75.0} & \textbf{95.0} & \textbf{90.0} &\textbf{85.0}\\
\bottomrule
\end{tabularx}

\label{tab:real1}
\end{table}

\subsection{Generalization Experiments}
In the context of learning robot manipulation policies,  OOD scenarios primarily refer to the emergence of visual information during the robot's inference process that was not present in the training dataset. These scenarios are often characterized by disturbances such as clutter, occlusion, and variations in lighting. In this section, we focus on the stability and robustness of our model compared to its baseline in the three generalization experimental tasks, as shown in Fig. \ref{fig:5}(b). Supplementary video demonstrations are available  at \href{https://GLUE666.github.io/}{https://GLUE666.github.io/}.

\textbf{Cluttered Background with Object Distractors:}

1) Motivation and Scene Setup: During the policy inference process, the introduction of cluttered, task-irrelevant objects can severely dilute and interfere with attention on visual representations. This causes the model's output to flatten, which manifests as hesitation or stopping in the robot's actions. However, the introduction of our global-local unified visual representation maps the OOD data back into a space that approximates the training dataset's distribution, which can effectively mitigate this issue. Therefore, we conducted an experiment by randomly placing a large number of irrelevant objects in the original task scene, while ensuring all other conditions remained exactly the same.

2) Experiment Results: The experimental results are presented in Table \ref{tab:ood1}. GLUE achieves high success rates on both tasks, whereas the baseline methods fail. On average, GLUE outperforms the strongest baseline by 60\% and surpasses GLUE-S by 30\%, this demonstrates that GLUE's unified global-local vision feature maintains robust performance in highly cluttered environments, unlike other policies that falter amidst such visual distractions.

\begin{table}[th]
\caption{
    Real-world experimental results in cluttered background scenes. In both tasks, GLUE achieves the highest performance, accomplishing tasks that the baselines fail to complete.
}
\centering
\small 
\renewcommand\arraystretch{1.3}
\setlength{\tabcolsep}{0mm}
\begin{tabularx}{\linewidth}{l *{9}{>{\centering\arraybackslash}X}}
\toprule
\begin{tabular}[t]{@{}c@{}} \footnotesize Method\end{tabular} &
\begin{tabular}[t]{@{}c@{}} \footnotesize Push  Button\end{tabular} &
\begin{tabular}[t]{@{}c@{}} \footnotesize Stack  Block\end{tabular} &
\begin{tabular}[t]{@{}c@{}} \footnotesize Average\end{tabular}
\\
\midrule

 DP \cite{dp}  & 0.0 & 0.0 & 0.0 \\
\rowcolor{mygrey}  ACT \cite{act}        & 0.0 & 0.0 & 0.0 \\
GLUE-S      & 50.0 & 10.0 & 30.0   \\

\midrule
    \noalign{\vskip -0.5ex}
    \rowcolor{mygrey} GLUE        & \textbf{80.0} & \textbf{40.0} & \textbf{60.0}  \\
\bottomrule
\end{tabularx}

\label{tab:ood1}
\end{table}

\textbf{Foreground Object Occlusion:}

1) Motivation and Scene Setup: Robots in real-world operational environments frequently face occlusions that can render a target object partially or completely invisible. This poses a significant challenge even for object-centric representation, as the visual representation becomes severely compromised. To validate our method's ability to operate robustly under such conditions, we introduce a large object to create severe occlusions of both the target object and the manipulator arm.

2) Experiment Results: The results are summarized in Table \ref{tab:ood2}. GLUE achieves high success rates on both tasks, while the baseline methods fail to complete them. On average, GLUE exceeds the strongest baseline by 70\% and outperforms GLUE-S by 25\%, highlighting its robust and reliable performance under severe foreground occlusion.

\begin{table}[th]
\caption{
    Real-world experimental results under foreground occlusion. GLUE demonstrates strong performance in both tasks, while the baselines fail to complete them.
}
\centering
\small 
\renewcommand\arraystretch{1.3}
\setlength{\tabcolsep}{0mm}
\begin{tabularx}{\linewidth}{l *{9}{>{\centering\arraybackslash}X}}
\toprule
\begin{tabular}[t]{@{}c@{}} \footnotesize Method\end{tabular} &
\begin{tabular}[t]{@{}c@{}} \footnotesize Push Button\end{tabular} &
\begin{tabular}[t]{@{}c@{}} \footnotesize  Place  Fruit\end{tabular} &

\begin{tabular}[t]{@{}c@{}} \footnotesize Average\end{tabular}
\\
\midrule

 DP \cite{dp}  & 0.0 & 0.0 & 0.0 \\
\rowcolor{mygrey}  ACT \cite{act}        & 0.0 & 0.0 & 0.0  \\
GLUE-S    & 40.0 & 50.0 & 45.0   \\

\midrule
    \noalign{\vskip -0.5ex}
    \rowcolor{mygrey} GLUE        & \textbf{70.0} & \textbf{70.0} & \textbf{70.0}  \\
\bottomrule
\end{tabularx}

\label{tab:ood2}
\end{table}

\textbf{Illumination Variations:}

1) Motivation and Scene Setup: Disturbances from changing light conditions are also very common in robotic manipulation environments. Variations in light can directly affect the colors of the environment and objects, which in turn disrupts the visual representation and can cause the model to fail. To test our model's robustness, we use a dynamic, colored party light (like a disco ball) to illuminate the experimental scene, thereby altering the lighting conditions while keeping all other factors constant.

2) Experiment Results: As illustrated in Table \ref{tab:ood3},GLUE achieved a success rate under lighting interference that was close to its performance in an interference-free environment, while all other baselines showed a significant drop in success rate. On average, our model achieves a 50\% improvement in success rate over the baseline, and outperforms the GLUE-s by 25\%. This demonstrates that our model can effectively handle the challenges posed by lighting variations.

\begin{table}[th]
\caption{
   Real-world experimental results under lighting variations. GLUE achieves the highest success rates in both tasks.
}
\centering
\small 
\renewcommand\arraystretch{1.3}
\setlength{\tabcolsep}{0mm}
\begin{tabularx}{\linewidth}{l *{9}{>{\centering\arraybackslash}X}}
\toprule
\begin{tabular}[t]{@{}c@{}} \footnotesize Method\end{tabular} &
\begin{tabular}[t]{@{}c@{}} \footnotesize  Place  Fruit\end{tabular} &
\begin{tabular}[t]{@{}c@{}} \footnotesize Fold Towel\end{tabular} &
\begin{tabular}[t]{@{}c@{}} \footnotesize Average\end{tabular}
\\
\midrule

 DP \cite{dp}  & 10.0 & 40.0 & 25.0 \\
\rowcolor{mygrey}  ACT \cite{act}        & 10.0 & 40.0 & 25.0  \\
GLUE-S     & 40.0 & 60.0 & 50.0   \\

\midrule
    \noalign{\vskip -0.5ex}
    \rowcolor{mygrey} GLUE        & \textbf{70.0} & \textbf{80.0} & \textbf{75.0}  \\
\bottomrule
\end{tabularx}

\label{tab:ood3}
\end{table}

\section{Conclusions and Limitations}
In this paper, we propose an imitation learning framework for the GLUE benchmark that unifies global and local visual representations. We first design a language-guided framework to automatically extract task-relevant, object-centric representations, using key object patches as our representation method. By fusing visual features from these key patches with global visual features, we obtain a highly robust visual representation that significantly improves the policy model's success rate. Furthermore, we enhance the imitation learning model's robustness against cluttered environments, occlusions, and lighting variations by mapping Out-of-Distribution (OOD) scenarios back to an approximate representation space of the original dataset. GLUE achieved results far superior to the other baseline models in simulation, real-world, and generalization scenarios, which strongly proves the effectiveness of our Global–Local Unified Encoding for Imitation Learning framework.

Our model still has a number of shortcomings. Because we use several large functional models like SAM and Co-tracker, our complete pipeline runs relatively slowly during practical deployment. Additionally, for long-horizon tasks, the tracking of our object-centric representations is under significant strain, meaning it is still possible for them to lose their targets.

\bibliographystyle{IEEEtran}
\bibliography{IROS}

\end{document}